\documentclass[conference]{IEEEtran}
\IEEEoverridecommandlockouts
\usepackage{cite}
\usepackage{amsmath,amssymb,amsfonts}
\usepackage{algorithmic}
\usepackage{graphicx}
\usepackage{textcomp}
\usepackage{xcolor}
\usepackage{booktabs}
\usepackage{adjustbox}
\usepackage{url} 
\usepackage{siunitx}
\usepackage[table]{xcolor}
\definecolor{lightyellow}{RGB}{255, 255, 204}
\usepackage{graphicx}
\usepackage{subcaption}

\def\BibTeX{{\rm B\kern-.05em{\sc i\kern-.025em b}\kern-.08em
    T\kern-.1667em\lower.7ex\hbox{E}\kern-.125emX}}
\begin{document}

\title{Gradient-based Model Shortcut Detection for Time Series Classification \\
\thanks{}
}

\author{
    \IEEEauthorblockN{
        Salomon Ibarra\IEEEauthorrefmark{2}, 
        Frida Cantu\IEEEauthorrefmark{2}, 
        Kaixiong Zhou\IEEEauthorrefmark{3}, 
        Li Zhang\IEEEauthorrefmark{2}
    }
    \IEEEauthorblockA{\IEEEauthorrefmark{2}Department of Computer Science, University of Texas Rio Grande Valley, USA\\
    \{salomon.ibarra01, frida.cantu02, li.zhang\}@utrgv.edu, kzhou22@ncsu.edu}
    \IEEEauthorblockA{\IEEEauthorrefmark{3}Department of Electrical and Computer Engineering, North Carolina State University, USA}
}

\maketitle

\begin{abstract}
Deep learning models have attracted lots of research attention in time series classification (TSC) task in the past two decades. Recently, deep neural networks (DNN) have surpassed classical distance-based methods and achieved state-of-the-art performance. Despite their promising performance, deep neural networks (DNNs) have been shown to rely on spurious correlations present in the training data, which can hinder generalization. For instance, a model might incorrectly associate the presence of grass with the label ``cat" if the training set have majority of cats lying in grassy backgrounds. However, the shortcut behavior of DNNs in time series remain under-explored. Most existing shortcut work are relying on external attributes such as gender, patients group, instead of focus on the internal bias behavior in time series models. 

In this paper, we take the first step to investigate and establish point-based shortcut learning behavior in deep learning time series classification. We further propose a simple detection method based on other class to detect shortcut occurs without relying on test data or clean training classes. We test our proposed method in UCR time series datasets. Source code is at \url{https://github.com/IvorySnake02/SAG.git}

\end{abstract}


\section{Introduction} 
Time series classification has been one of the most well-researched tasks due to the wide application~\cite{dau2019ucr, zhang2020tapnet,bagnall2017great, guo2020multivariate}. Recent years, deep neural networks have shown promising performance compared to traditional distance based method~\cite{dempster2020rocket,zha2022towards}. Meanwhile, recent studies show deep learning models often rely on simple or superficial patterns spuriously correlated in the data that are not causally or semantically related to the true task~\cite{geirhos2018imagenet,lai2021machine, zhang2018examining}. This phenomenon is also known as \textit{shortcut learning}~\cite{geirhos2020shortcut}. For example, a model trained on animal classification are learning correlation on the background rather than animal features~\cite{beery2018recognition, geirhos2020shortcut}. Due to the ubiquitous existence in time series in different domains such as health~\cite{wang2022systematic}, finance~\cite{skabar2013direction}, and manufacturing~\cite{shi2023lstm}, shortcut learning problems could significantly hurt generalization ability and lead to serious issue in real-world mission-critical tasks. 
\begin{figure}[t]
    \centering
    \includegraphics[width=1\linewidth]{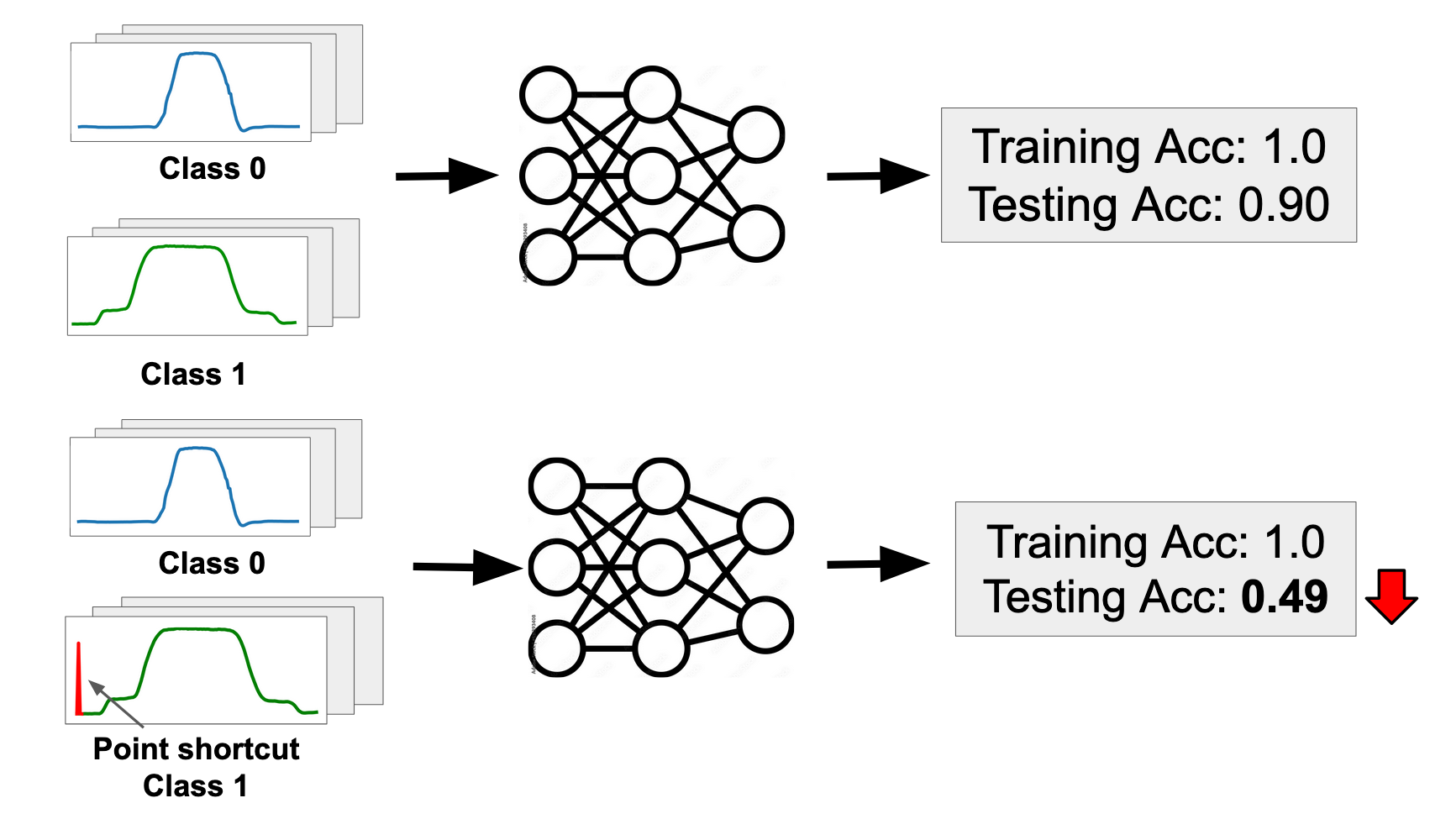}
    \caption{Training and testing accuracies of ResNet\cite{he2016deep} models on the GunPoint\cite{UCRArchive} dataset (top) and on the modified GunPoint dataset with a point shortcut feature added to all class 1 training samples (bottom) after 100 epochs.}
    \label{fig:motivation}
\end{figure}

Despite numerous studies that examined shortcut behavior of DNNs for image and text data~\cite{beery2018recognition, du2023shortcut, geirhos2020shortcut}, the characteristics of shortcut behavior on time series data are largely underexplored. Most existing work focus on image data on image specific features such as texture and gray level~\cite{hermann2023foundations}. Existing shortcut learning in time series often relies on external attributes such as age group or gender, leading to biased models and fairness concerns in health domain~\cite{brown2023detecting}. For shortcut detection, existing shortcut detection approaches include data augmentation based on some known features such as image background~\cite{choi2019can}, or measuring statistical parity~\cite{brown2023detecting}. However, external attributes are often unavailable in practical time series settings, it is unclear how to identify shortcut features within the time series itself.

We demonstrate deep neural networks could be easily mislead by learning with a simple experiment on ResNet18 with GunPoint data in UCR Time Series Archive~\cite{UCRArchive}. As shown in Figure~\ref{fig:motivation} we manually inject a small spike on a fixed position in one class of each instance training data of the postive class, and re-train the model, and the accuracy dropped from 90\% to 49\%. This means that deep learning model prefers to learn \textit{shortcut} --using a spike to make decision rather than utilizing meaningful contextual features. 

In this paper, we take the first step to investigate and establish the point shortcut problem. We show it widely exist in deep learning models in time series. To address the point shortcut problem, we propose a simple gradient aggregation score based on the input after the model training to detect class-based shortcut, without requiring external attributes or knowledge about testing data. We show the effectiveness of our method in detecting point shortcut with UCR time series datasets. 

\section{Related Work} 
Existing studies show shortcut learning exists in image and natural language data. Existing work \cite{beery2018recognition}\cite{geirhos2020shortcut} shows that deep learning models are biased toward learning shortcuts in backgrounds, colors, textures, etc. In natural language domain, DNN models are highly dependent on unintended repeating features to
make predictions, and simple surface-level features and negation words~\cite{lai2021machine}\cite{niven2019probing} instead of actual semantic meaning. 

In time series domain, shortcut learning behavior of deep learning models are largely underexplored. Most existing work focus on unfair performance, instead of identify model bias inherently existing in time series model during or after training. Although spectral bias is found to exist in time series transformers\cite{ackaah2023exploring} on long range forecasting task, often the model bias is formulated into a disparity issue on external predefined group such as gender and age, or specific type of disease detection rate on  classes~\cite{bhanot2021problem, he2023learning, brown2023detecting}. To the best of our knowledge, there is no existing work attempting to demonstrate the point-based shortcuts existing in time series classification models, and there is no direct similar work detecting point-based shortcut solely based on training data and model training in time series. 

\section{Problem Statement} \label{sec:problem}

In this paper, we study the problem of \textit{shortcut detection for time series classification models}. Let $X$ denote a collection of $n$ time series instances of length $m$, and $Y\in C^n$ denote their class labels where $C = \{0, 1\}$. We would like to train a neural network as a classification model such that $f(x_i; \theta) = \hat{y}$ where $\hat{y}$ is the predicted class of instance $x_i$. 

We define a \textit{shortcut feature} $x^s$ as a set of unintended features that is highly correlated with the label $y$ that the neural network $f_\theta^s$ can achieve high accuracy in training set, but cannot generalize to testing set. The task of \textit{shortcut detection} aims to detect shortcut feature in training data $X^{train}$ that mislead models, thus enhancing the model reliability.   



\begin{figure}[htbp]
    \centering
    \begin{subfigure}[b]{0.9\linewidth}
        \includegraphics[width=\linewidth]{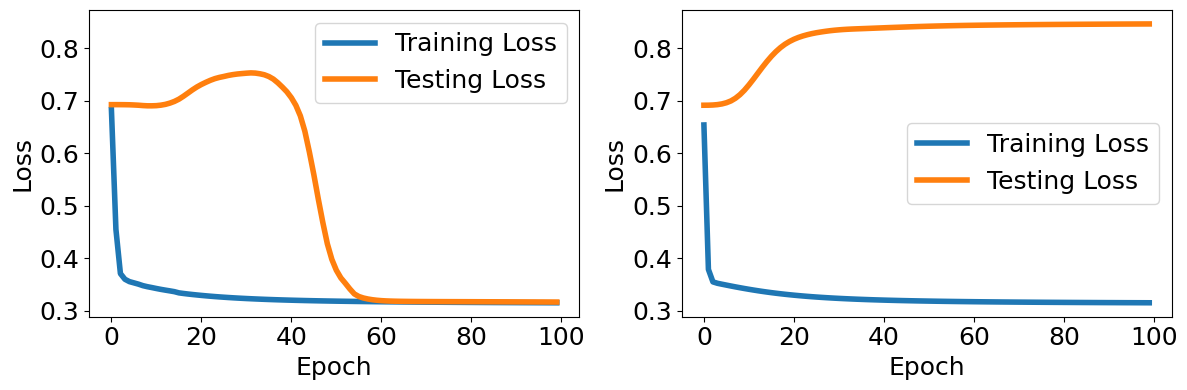}
        \caption{Coffee: Before vs. After adding point shortcut}
        \label{fig:coffee}
    \end{subfigure}
    \hfill
    \begin{subfigure}[b]{0.9\linewidth}
        \includegraphics[width=\linewidth]{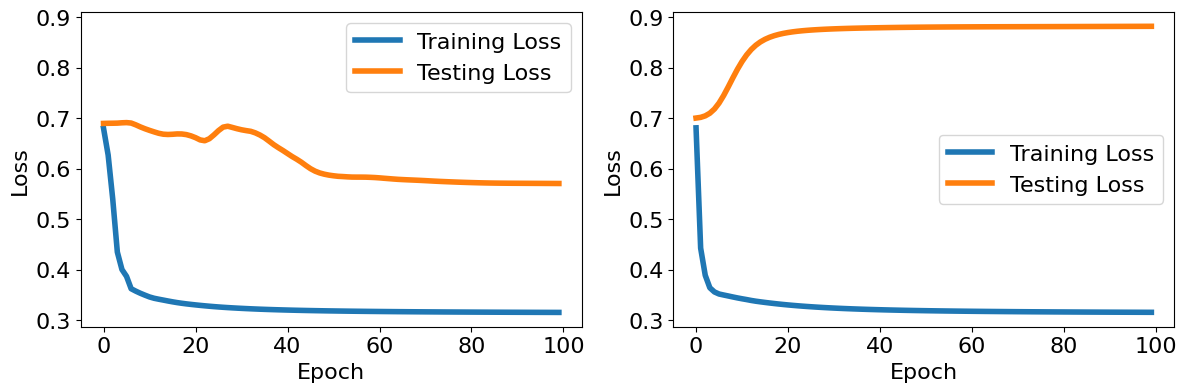}
        \caption{WormsTwoClass: Before vs. After adding point shortcut}
        \label{fig:worms}
    \end{subfigure}
    \hfill
    \begin{subfigure}[b]{0.9\linewidth}
        \includegraphics[width=\linewidth]{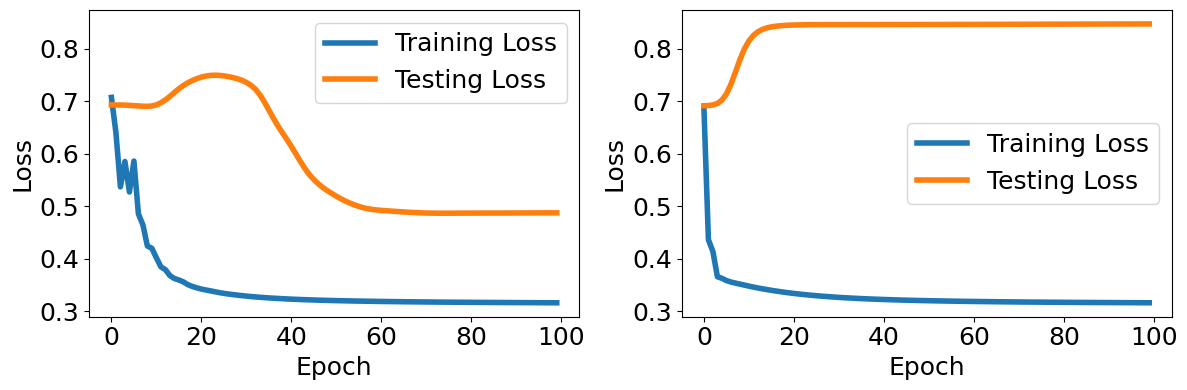}
        \caption{Yoga: Before vs. After adding point shortcut}
        \label{fig:yoga}
    \end{subfigure}
    \caption{Resulting loss curves for ResNet18 model on Coffee, WormsTwoClass, and Yoga. Left: original datasets. Right: modified datasets with point shortcut features added to training only.}
    \label{fig:combined}
\end{figure}

\section{Preliminary Observations}\label{sec:prelim}

To test the impact of point shortcut, we conduct a comprehensive point-based injection in all UCR classification archive \cite{UCRArchive} with instance length less than 1000. Out of 40 datasets qualified, at least 24 datasets are clearly impacted by the point-shortcut based on training with Resnet. The detailed accuracy and loss function plots comparison can be found in our supporting webpage~\footnote{\url{https://github.com/IvorySnake02/SAG.git}}.
 We will discuss more details about the experiment setting in Section~\ref{sec:exp}.

Figure~\ref{fig:combined} shows the comparison of shortcut impact of training and testing loss over epochs in three of the 24 datasets. We first trained ResNet~\cite{he2016deep} on the original data, then added a simple point shortcut feature in training data only, and re-train the model. A single point-based shortcut can cause the model to latch onto unintended features during the early training epochs, resulting in poor generalization on unseen data, as evidenced by the significantly elevated testing loss (orange curves) in the right-hand side panels in Figure~\ref{fig:combined}, which shows that in time series data, even without carefully optimization, simple and strongly correlation with labels can easily mislead deep learning models.

\section{Proposed Method}
In this section, we introduce our proposed method Shortcut Aggregate Gradient score (SAG) that takes advantage of shortcut models unique input gradient to determine the existence of shortcut features within our training data.  

\subsection{Shortcut Aggregate Gradient Score}
Since shortcuts often manifest as simple correlations that models can easily capture, we design a new shortcut score by aggregating input gradients and examining whether individual points show abnormally high average gradient values. 
Specifically, we compute our \textit{Point-shortcut Score} as follows:  
\begin{equation}
    \delta_{t,c} = \left|\frac{1}{n_c} \sum_{{x_i \in c}} \frac{\partial L_{CE}}{\partial x_{it}}, \right|,
    \label{eqn:mag}
\end{equation}
where $t$ is the time stamp on a time series sample, $L_{CE}$ is cross-entropy loss for model $f_{\theta}$. $\frac{\partial L_{CE}}{\partial x_i}$ denotes the gradient of the pretrained model with respect to $i$-th sample. 

Finally, we define the proposed \textit{Shortcut Aggregation Gradient (SAG)} score as: 
\begin{equation}
    \text{SAG}(c) = \frac{max_t \delta_{t,c}}{\sum_t \delta_{t,c}},
    \label{eqn:sags_score}
\end{equation}

\noindent where $\delta$ is the class-wise gradient importance score for class $c$. 

The SAG score measures the extent to which a single class dominates the gradient importance distribution. 

We use a sensitivity threshold $\epsilon$ to determine whether the pre-trained model has a shortcut class based on the proposed SAG score to determine if a dataset or a class has shortcut. We define the shortcut detection function as:  
\begin{equation}
    D(X) = 
    \begin{cases} 
        1, & \text{if } \max_c \text{SAG}(c) > \epsilon, \\
        0, & \text{otherwise},
    \end{cases}
    \label{eqn:shortcut_detect}
\end{equation}
where $D(X)=1$ indicates that shortcut is detected in the model, and $D(X)=0$ indicates no significant shortcut is present.
In other words, if any class of the dataset has the SAG score is greater than $\epsilon$, we will report that we locate a model shortcut and which class it was detected.

\section{Experiment}\label{sec:exp}
In this section, we will describe the experiment setting and evaluation, and show quantative results of our proposed method. 
\subsection{Experiment setting}
\subsubsection{Datasets}
We use all the datasets from the UCR archive~\cite{UCRArchive} with two classes, fewer than 1,000 training samples, and a time series length less than or equal to 1,000 to perform the experiment. To test the effect of point shortcut, for each data, we injected a point shortcut at the first point in all the samples in the positive class only, while keeping the negative class in the training data as well as testing data remain the same. We validate the shortcut by examining the training and testing training loss compare with the original (as shown in Figure 2). Out of 40 data qualified, 24 datasets showed a significant sign of shortcut based on the testing loss compared with training as shown in Table~\ref{tab:resnet_results}. For shortcut detection, we use all 24 datasets to evaluate our results.
\begin{figure}[htbp]
    \centering
    \begin{subfigure}[b]{0.45\linewidth}
        \centering
        \includegraphics[width=\linewidth]{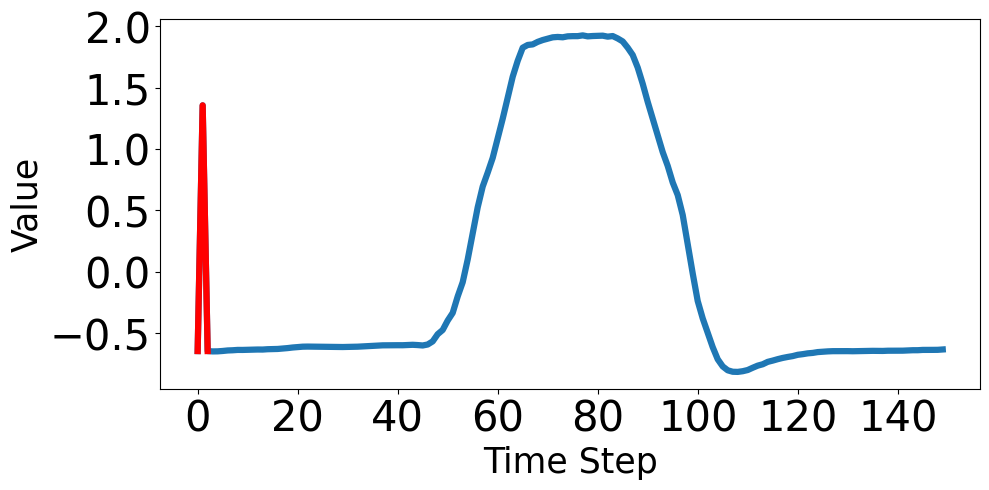}
        \caption{Gunpoint}
        \label{fig:gunpoint_class}
    \end{subfigure}
    \hfill
    \begin{subfigure}[b]{0.45\linewidth}
        \centering
        \includegraphics[width=\linewidth]{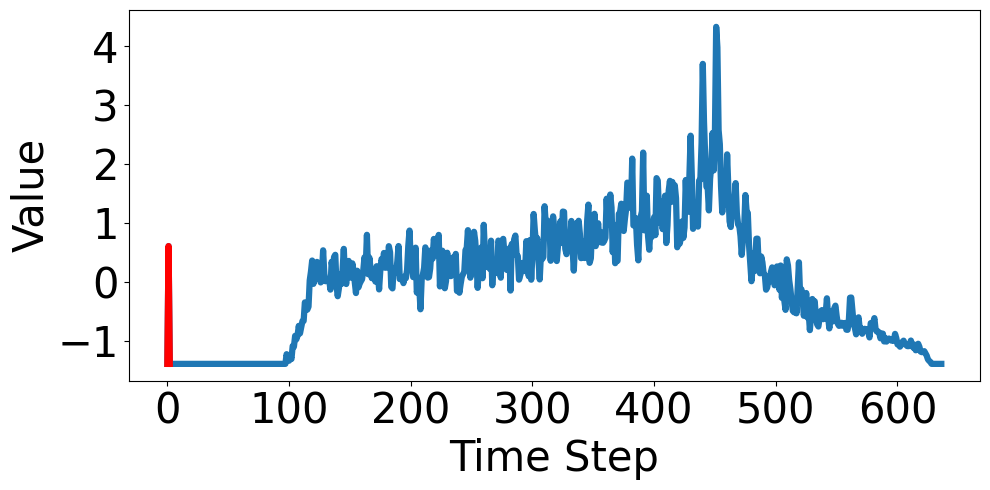}
        \caption{Lightning}
        \label{fig:lightning_class}
    \end{subfigure}

    \vskip\baselineskip

    \begin{subfigure}[b]{0.45\linewidth}
        \centering
        \includegraphics[width=\linewidth]{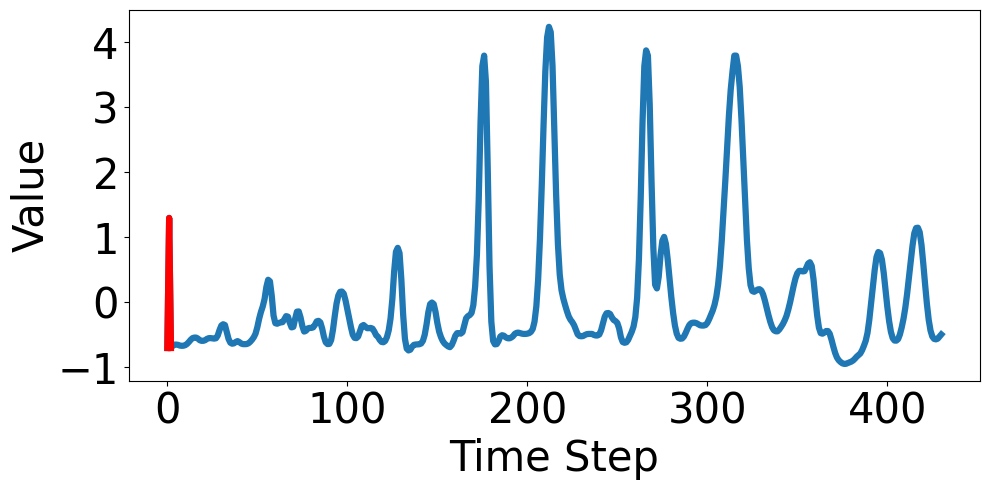}
        \caption{Ham}
        \label{fig:ham_class}
    \end{subfigure}
    \hfill
    \begin{subfigure}[b]{0.45\linewidth}
        \centering
        \includegraphics[width=\linewidth]{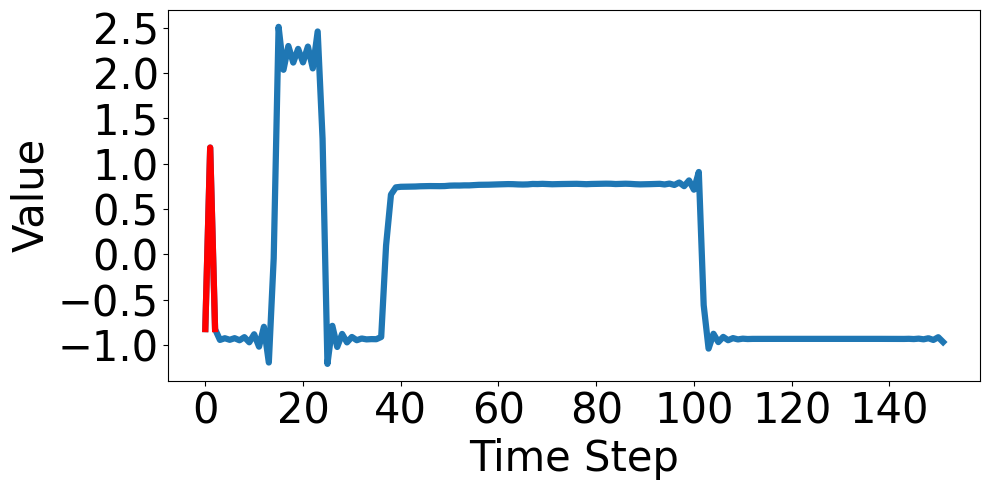}
        \caption{Wafer}
        \label{fig:wafer_class}
    \end{subfigure}

    \caption{Visualizations for datasets that are modified to include a point shortcut.}
    \label{fig:all_classes}
\end{figure}

\subsubsection{Base Models} 
 We used \textbf{ResNet18}\cite{he2016deep} as our base model, since ResNet has shown good performance in time series classification.  follow~\cite{barreda2024cosco,xi2024efficient} parameter with Adam optimizer~\cite{kingma2014adam} with a learning rate of .001 and being trained for up to 100 epochs. The experiment was carried out on Google Colab\footnote{\url{https://colab.research.google.com/}} using a T4 16GB GPU.


\subsection{Evaluation}

To evaluate the effectiveness of our method, we employ two metrics: \textit{Class Detection Accuracy} and \textit{Dataset Detection Accuracy}. Particularly, Class Detection Accuracy evaluates shortcut identification at the class level, 
whereas Dataset Detection Accuracy evaluates shortcut identification at the dataset level. 

\begin{equation}
    \text{Class Detection Accuracy} = \frac{\text{correct shortcut class detected}}{\text{number of datasets}}
\end{equation}\label{eqn:metricA}
\begin{equation}
    \text{Dataset Detection Accuracy} = \frac{\text{correct shortcut dataset detected}}{\text{number of datasets}}
\end{equation}\label{eqn:metricB}

Together, these two metrics provide complementary perspectives on the reliability of the proposed SAG Score.

\begin{table*}[htbp]
\centering
\caption{Per-class Short Aggregate Gradient scores under regular and shortcut training for ResNet18}
\label{tab:resnet_results}
\begin{tabular}{lcccccccc}
\toprule
\textbf{Dataset} & \textbf{Train Size} & \textbf{Length} & \textbf{Classes} & \textbf{Type} &
\multicolumn{2}{c}{\textbf{Regular Training}} & \multicolumn{2}{c}{\textbf{Shortcut Training}} \\
\cmidrule(lr){6-7} \cmidrule(lr){8-9}
& & & & & \textbf{Class 0} & \textbf{Class 1} & \textbf{Class 0} & \textbf{Class 1} \\
\midrule
BeetleFly & 20 & 512 & 2 & IMAGE & 0.0487 & 0.0663 & 0.0439 & \textbf{0.1995} \\
BirdChicken & 20 & 512 & 2 & IMAGE & 0.0305 & 0.0211 & 0.0788 & \textbf{0.2588} \\
Coffee & 28 & 286 & 2 & SPECTRO & 0.0324 & 0.0313 & 0.0318 & 0.1123 \\
DistalPhalanxOutlineCorrect & 600 & 80 & 2 & IMAGE & 0.0668 & 0.0640 & 0.0489 & \textbf{0.2006} \\
DodgerLoopWeekend & 20 & 288 & 2 & SENSOR & 0.0256 & 0.0666 & 0.0284 & 0.0247 \\
ECG200 & 100 & 96 & 2 & ECG & 0.0579 & 0.0642 & 0.1430 & 0.1030 \\
ECGFiveDays & 23 & 136 & 2 & ECG & 0.0558 & 0.0536 & 0.0829 & \textbf{0.3432} \\
Epilepsy2 & 80 & 178 & 2 & EEG & 0.0385 & 0.0405 & 0.0461 & 0.0724 \\
FreezerRegularTrain & 150 & 301 & 2 & DEVICE & 0.1036 & 0.0974 & 0.1447 & \textbf{0.3209} \\
GunPoint & 50 & 150 & 2 & HAR & 0.0350 & 0.0356 & 0.0469 & \textbf{0.2176} \\
Ham & 109 & 431 & 2 & SPECTRO & 0.0260 & 0.0168 & 0.1860 & \textbf{0.2935} \\
ItalyPowerDemand & 67 & 24 & 2 & SENSOR & 0.1457 & 0.0992 & 0.1527 & \textbf{0.2511} \\
Lightning2 & 60 & 637 & 2 & SENSOR & 0.0155 & 0.0152 & 0.0507 & \textbf{0.1749} \\
MiddlePhalanxOutlineCorrect & 600 & 80 & 2 & IMAGE & 0.0598 & 0.0465 & 0.0550 & \textbf{0.2709} \\
MoteStrain & 20 & 84 & 2 & SENSOR & 0.0633 & 0.0816 & 0.0552 & 0.1363 \\
PowerCons & 180 & 144 & 2 & DEVICE & 0.0557 & 0.0393 & 0.1644 & \textbf{0.2536} \\
SharePriceIncrease & 965 & 60 & 2 & FINANCIAL & 0.0487 & 0.0559 & 0.2566 & \textbf{0.1994} \\
SonyAIBORobotSurface2 & 27 & 65 & 2 & SENSOR & 0.0569 & 0.0509 & 0.0605 & \textbf{0.1762} \\
ToeSegmentation1 & 40 & 277 & 2 & MOTION & 0.0212 & 0.0247 & 0.1015 & \textbf{0.1815} \\
ToeSegmentation2 & 36 & 343 & 2 & MOTION & 0.0371 & 0.0288 & 0.0392 & \textbf{0.2230} \\
TwoLeadECG & 23 & 82 & 2 & ECG & 0.0694 & 0.0706 & 0.0643 & \textbf{0.2187} \\
Wafer & 1000 & 152 & 2 & SENSOR & 0.0579 & 0.0634 & 0.1441 & \textbf{0.1980} \\
WormsTwoClass & 181 & 900 & 2 & MOTION & 0.0112 & 0.0094 & 0.0496 & \textbf{0.2996} \\
Yoga & 300 & 426 & 2 & IMAGE & 0.0247 & 0.0252 & 0.0258 & \textbf{0.2494} \\
\midrule
$\epsilon = 0.15$ & & & & & $< \epsilon$ & $< \epsilon$ & $< \epsilon$ & $> \epsilon$ \\
\textbf{Correct Class Predictions (out of 24)} & & & & & 24 & 24 & 20 & 19 \\
\textbf{Class Detection Accuracy} & & & & & 1.000 & 1.000 & 0.833 & 0.792 \\
\textbf{Dataset Detection Accuracy} & & & & & \multicolumn{2}{c}{1.000} & \multicolumn{2}{c}{0.792} \\
\bottomrule
\end{tabular}
\end{table*}

\begin{figure}
    \centering
    \includegraphics[width=0.9\linewidth]{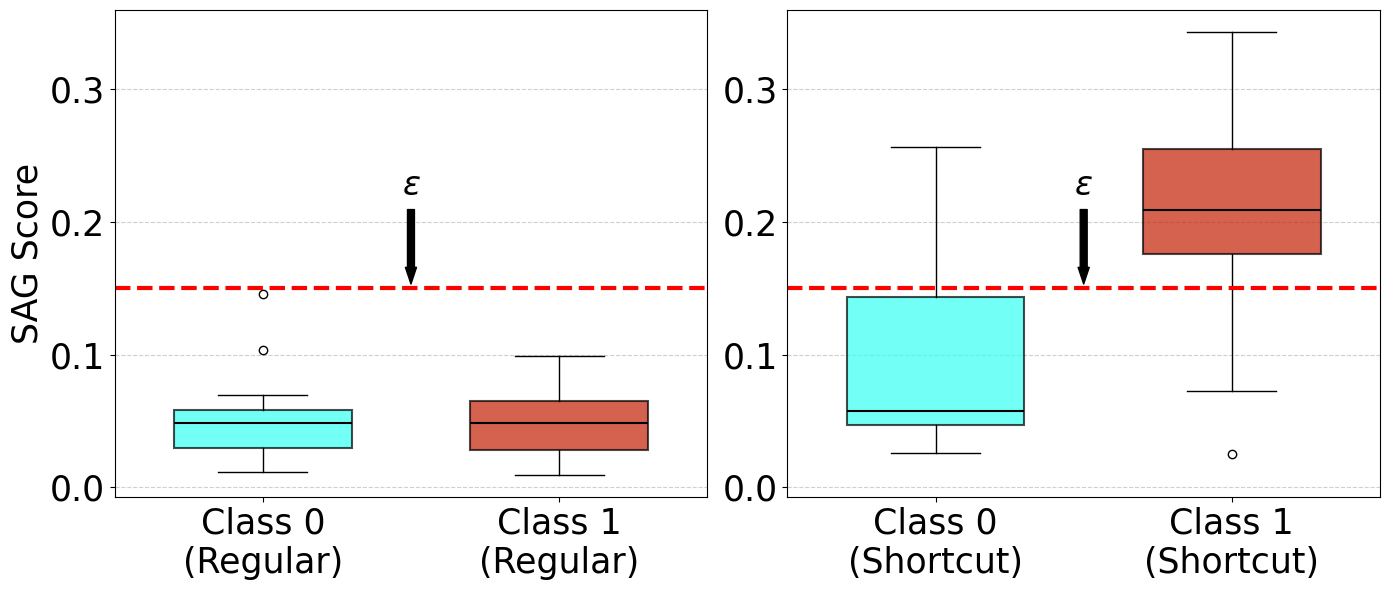}
    \caption{SAG scores for regular dataset (left) and shortcut dataset (right) showing the effectiveness of parameter $\epsilon$ for filtering shortcut-affected samples.}
    \label{fig:boxplot}
\end{figure}

\subsection{Result}
We perform a similar experiment to the one we performed in section \ref{sec:prelim}. We train our base model twice once with the regular data and then again after the data was modified. Then after training evaluate our proposed methods performance using our two proposed metrics. Table \ref{tab:resnet_results} shows the results of our experiment. We used $\epsilon = 0.15$ to filter for positive results, wins for true positive shortcut classes detected are in bold. Our proposed method had a success rate 100\% with no false positives effectively ignoring irrelevant data sets. Our method also has an 83\% accuracy detecting non-shortcut classes with in datasets that contain one shortcut class. Furthermore, our method has a success rate of 79\% in correctly detecting shortcut classes and shortcut data sets. In addition, we can see our parameter $\epsilon$ succeed at filtering in Figure~\ref{fig:boxplot}, which shows that, on average, $\epsilon$ is capable of distinguishing shortcut-affected samples (Class 1 in the shortcut datasets) from regular samples while leaving unaffected classes below $\epsilon$. Given these experiment results, we can conclude that our proposed method is highly precise and conservative with false positive rates and a near 4 in 5 chance of successfully detecting a model shortcut.
\subsection{Visualizing GunPoint Point-shortcut Score}
\begin{figure}[htbp]
    \centering    \includegraphics[width=0.5\textwidth]{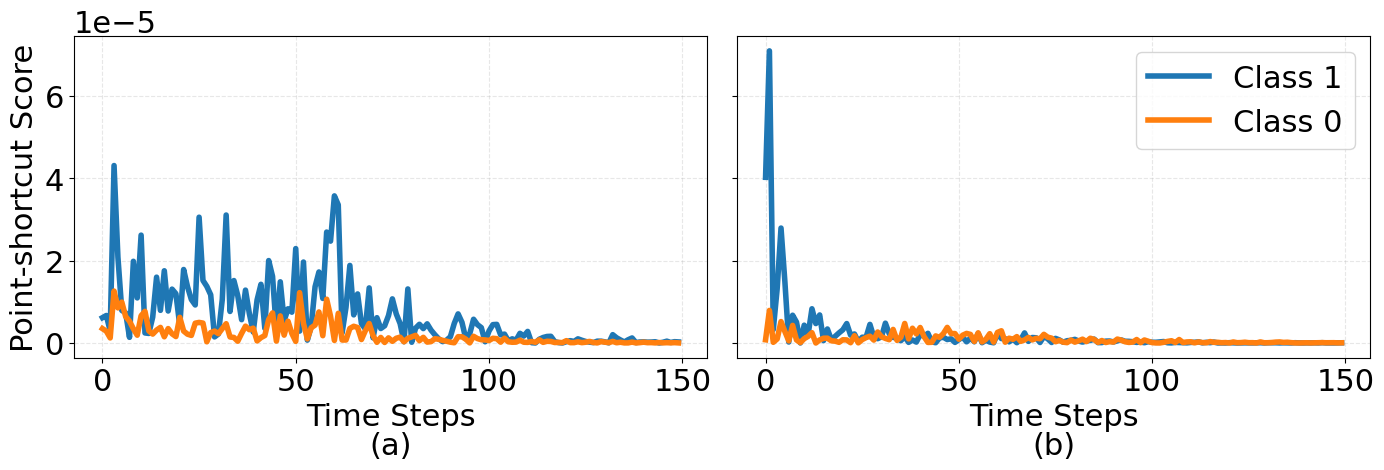}
    \caption{Visualizing the point-shortcut score on ResNet18 trained on (a) original VS. (b) point-shortcut injected to GunPoint data~\cite{UCRArchive}. }\label{fig:gunpoint_grad}
\end{figure}

We show the effectiveness of shortcut score $\delta_{t,c}$ by visualizing Gunpoint data. When training with orginal data, the shortcut score based on the input gradient is pretty evenly spreaded out as shown in Figure \ref{fig:gunpoint_grad} (a). Class 1 has a slight higher magnitude on the average input gradient. When training with shortcut injected injected at first point of the sequence, we can observe that the average input gradient is simply concentrated on a single point, which is around the location of shortcut point. That indicate the simple unintended relationship is captured by deep learning model, and preventing the model to learn the semantic information in other locations in class 1, while class 0 is not highlighted. The result shows that our proposed score is able to detect the shortcut class and align with our intuition of model shortcut behavior. 

\begin{figure}[htbp]
    \centering
    \begin{subfigure}[b]{0.30\linewidth}
        \centering
        \includegraphics[width=\linewidth]{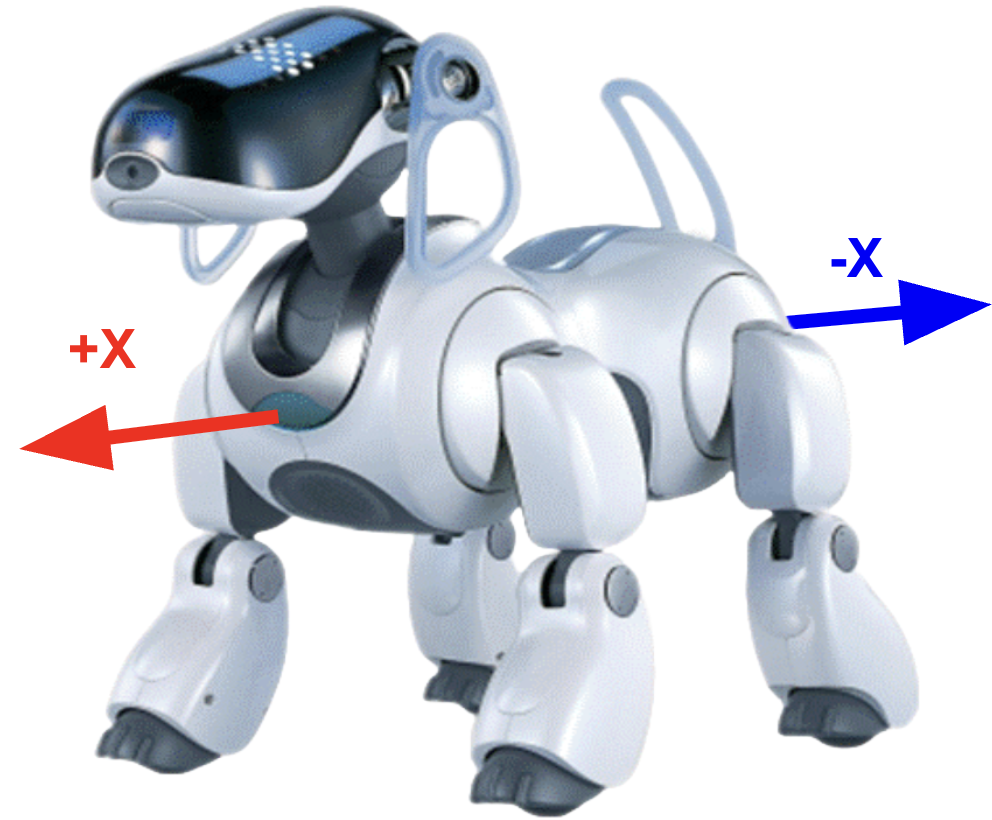}
        \caption{Robot}
        \label{fig:robot-dog}
    \end{subfigure}
    \hfill
    \begin{subfigure}[b]{0.60\linewidth}
        \centering
        \includegraphics[width=\linewidth]{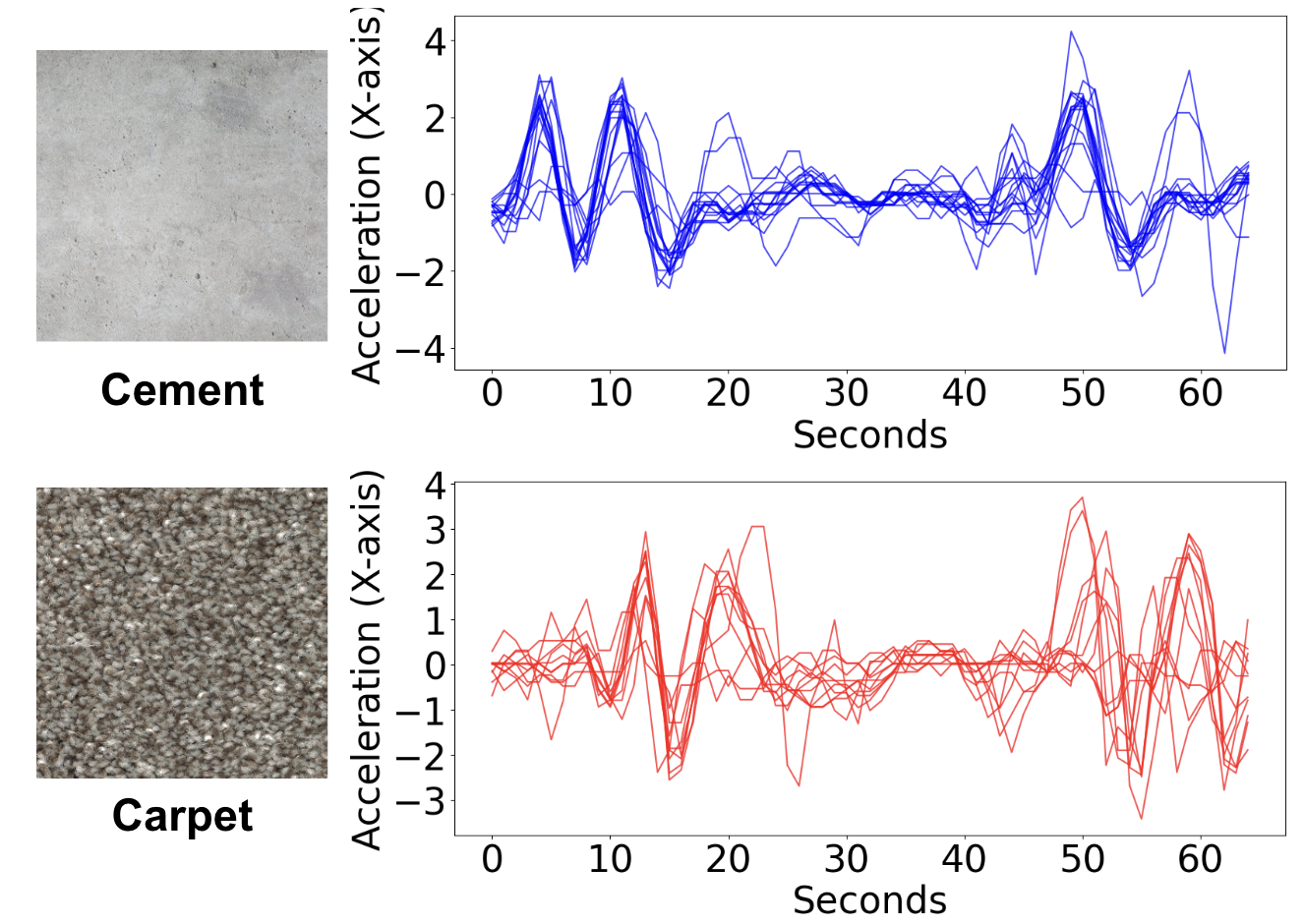}
        \caption{Classes}
        \label{fig:surfaces}
    \end{subfigure}
    \caption{(a) Sony AIBO Robot ERS210a \cite{mueen2011logical}, with positive and negative x-axis labeled. (b) Class 0: Cement and Class 1: Carpet}
    \label{fig:case-study-sony}
\end{figure}

\vspace{-1mm}
\subsection{Case Study: SonyAI Robot Surface}
In this case study, we use the SonyAIBORobotSurface2 dataset to show that our method can effectively capture the shortcut in the data injected with this dataset using base model ResNet18. We obtained SonyAIBORobotSurface2 dataset from the UCR Time Series Classification archive. The dataset comes from Vail et. al. \cite{VAIL2004822}, where a Sony AIBO ERS210a, a four-legged robot resembling a dog as shown in Figure \ref{fig:robot-dog}, was made to walk over different surfaces such as cement and carpet to optimize robot calibration in different environments. The Sony robot had an internal accelerometer sensor, which captured the robot's acceleration in three dimensions (x,y,z). The SonyAIBORobotSurface2 dataset only contains the x-axis readings where x is the direction of the robot going forward and backward as we labeled in Figure \ref{fig:robot-dog}. The data set has 27 training samples with a time series length of 65 at a sampling rate of 65 Hz. Figure \ref{fig:surfaces} shows that the datasets two classes: class 0 where the robot is walking on cement surface and class 1 robot walking on a carpeted surface.
\begin{figure}
    \centering
    \includegraphics[width=1\linewidth]{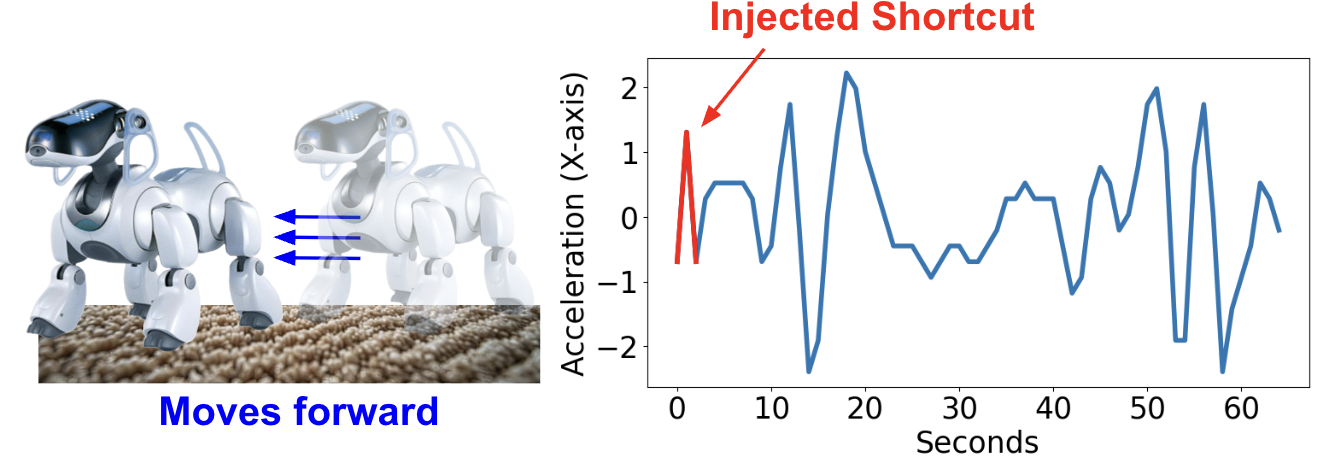}
    \caption{Sample from class 1 where shortcut was injected in the beginning with illustration on the left of the Sony robot dog \cite{mueen2011logical}, showing the corresponding action.}
    \label{fig:spiked-sony-sample}
\end{figure}

\begin{figure}[htbp]
    \centering

    \begin{subfigure}[b]{0.45\linewidth}
        \centering
        \includegraphics[width=\linewidth]{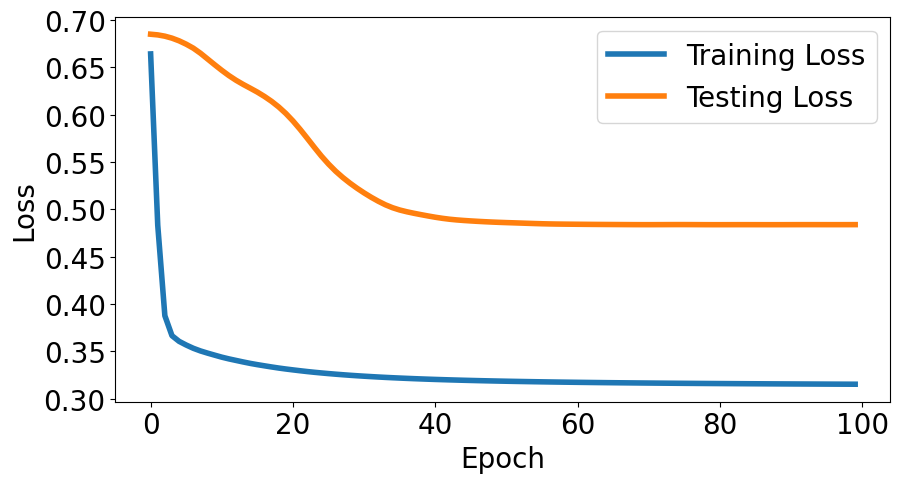}
        \caption{Loss of Regular data}
        \label{fig:SonyRobotLoss}
    \end{subfigure}
    \hfill
    \begin{subfigure}[b]{0.45\linewidth}
        \centering
        \includegraphics[width=\linewidth]{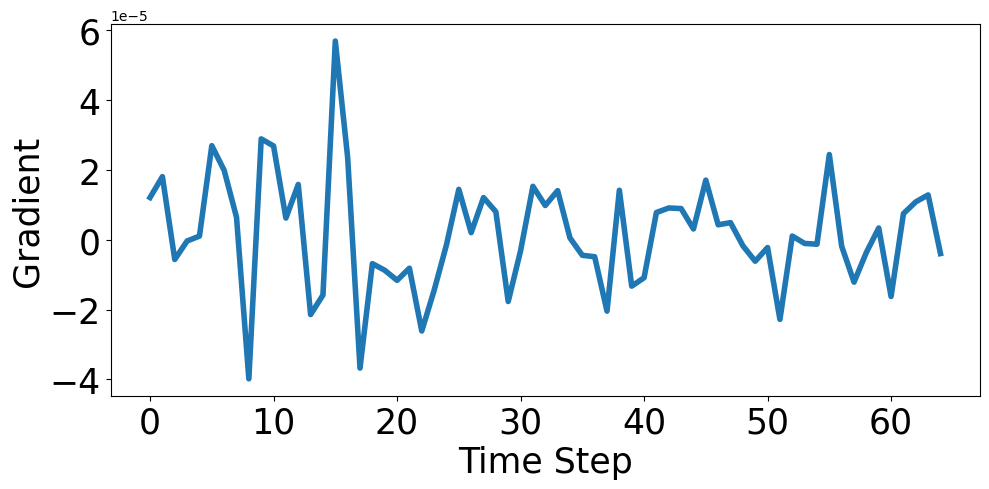}
        \caption{Gradient of Regular data}
        \label{fig:SonyRobotGrad}
    \end{subfigure}
    
    \vskip\baselineskip
    
     \begin{subfigure}[b]{0.45\linewidth}
        \centering
        \includegraphics[width=\linewidth]{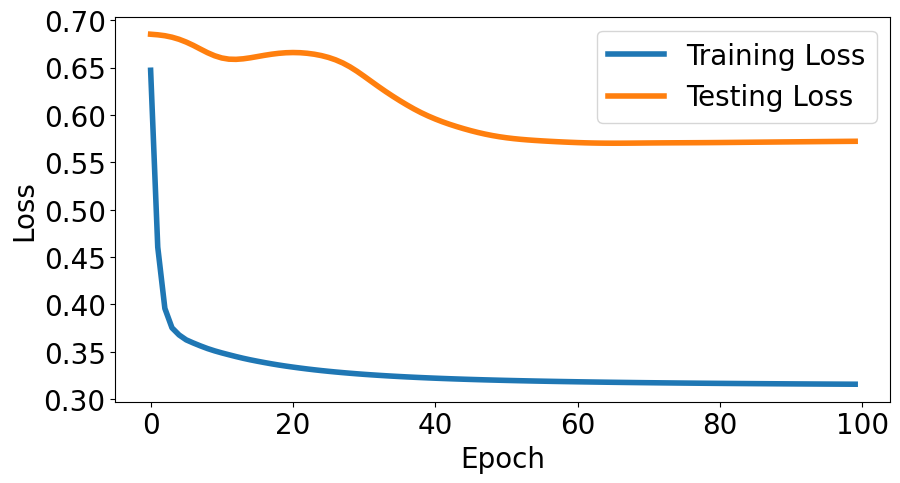}
        \caption{Loss After Shortcut}
        \label{fig:SonyRobotLoss_spike}
    \end{subfigure}
    \hfill
    \begin{subfigure}[b]{0.45\linewidth}
        \centering
        \includegraphics[width=\linewidth]{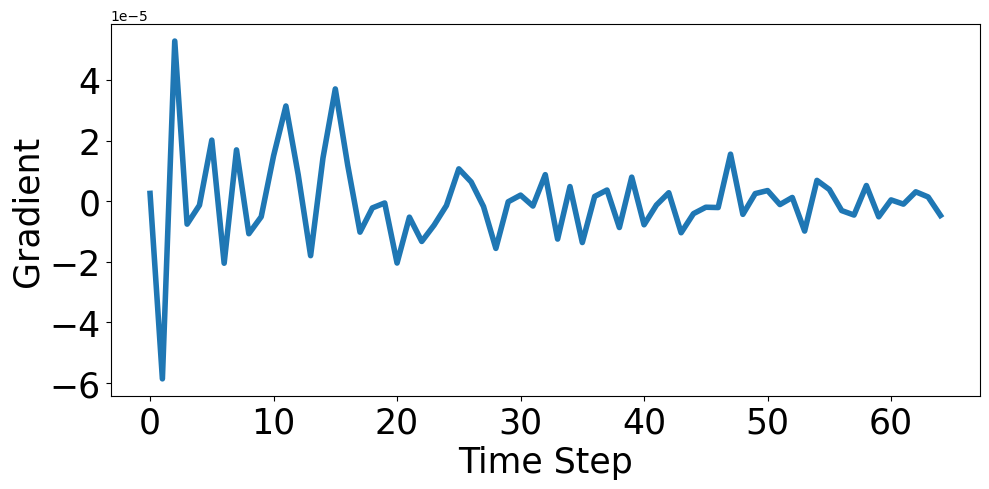}
        \caption{Gradient After Shortcut}
        \label{fig:SonyRobotGrad_spike}
    \end{subfigure}
    \caption{SonyAIBORobotSurface2 loss and gradient plots before and after injecting the shortcut, using ResNet18.}
    \label{fig:Sony-robot-plots}
\end{figure}
In our experiment, we want to see how the injection of a shortcut affects the performance of the model. In Figure \ref{fig:spiked-sony-sample}, we injected a spike at the beginning of each class 1 sample to serve as a shortcut (red), which is a simulated move towards forward move on a carpet surface, as shown in the illustration (left). The insertion of the shortcut into the beginning of all the samples in class 1 distracts the model from the rest of the features in the data, which we can see in Figure \ref{fig:Sony-robot-plots}. Furthermore, Figure \ref{fig:SonyRobotLoss_spike} shows that the test loss after injecting the shortcut dips at around 20 epochs and plateaus for the rest of the 100 epochs compared to \ref{fig:SonyRobotLoss} where the model continues learning before plateauing at around 30 epochs. Figure \ref{fig:SonyRobotGrad} shows the mean input gradient between class 0 and class 1 before injecting the shortcut, and compared to \ref{fig:SonyRobotGrad_spike}, the range of values for the gradient is consistent, which means that the model is learning the features in the data. In \ref{fig:SonyRobotGrad_spike}, the largest range of values is in the beginning where the model picked up the shortcut and then the gradient values ranges closer to zero. Based on these results, we observed that the injection of the shortcut into the SonyAIBORobotSurface2 dataset affected the performance of the ResNet18 model.
\section{Conclusion}
In this paper, we confirm the phenomenon of point-based shortcuts in time series data. We perform a preliminary observation and report the characteristics that a time series classification deep neural network that learned a shortcut exhibits. We also proposed a score and threshold that rely on the unique input gradient of a model that learned shortcut to determine the presence of a shortcut or shortcut class in a dataset. Then we perform a case study on Sony AI robot from UCR time series classification archive to evaluate how detrimental a shortcut is to a models performance.
\section*{Acknowledgment}

This work was supported in part by the National Science Foundation under Grants (CNS-2431514, IIS-2434916, CNS-2431516). The computational resource is supported by Google Cloud Platform (GCP) credits. 

\bibliographystyle{plain}
\bibliography{shortcut}
\end{document}